\documentclass[runningheads]{llncs}

\usepackage{eccv}

\usepackage{eccvabbrv}
\usepackage{soul}
\usepackage{graphicx}
\usepackage{booktabs}
\usepackage{booktabs}       % professional-quality tables
\usepackage{multirow}
\usepackage{epsfig}
\usepackage{graphicx}
\usepackage{amsmath}
\usepackage{amssymb}
\usepackage{booktabs}
\usepackage{bbm}
\usepackage{wrapfig}
\usepackage{colortbl}
\usepackage{tabulary}
\usepackage{etoolbox}
\usepackage{overpic}

\usepackage{arydshln}
\usepackage{listings}
\usepackage{tabularx}
\definecolor{codegreen}{rgb}{0,0.5,0}
\definecolor{codeblue}{rgb}{0.25,0.5,0.5}
\definecolor{codegray}{rgb}{0.6,0.6,0.6}
\definecolor{eccvblue}{rgb}{0.12,0.49,0.85}

\usepackage[ruled,linesnumbered]{algorithm2e}
\usepackage{listings}
\definecolor{codegreen}{rgb}{0,0.5,0}
\definecolor{codeblue}{rgb}{0.25,0.5,0.5}
\definecolor{codegray}{rgb}{0.6,0.6,0.6}

\usepackage[accsupp]{axessibility}  % Improves PDF readability for those with disabilities.

\newtheorem{prop}{Corollary}

\newcommand*{\bbone}{\boldsymbol{1}}
\usepackage{hyperref}
\definecolor{ao}{rgb}{0.0, 0.0, 1.0}
\definecolor{airforceblue}{rgb}{0.36, 0.54, 0.66}
\definecolor{ceruleanblue}{rgb}{0.16, 0.32, 0.75}
\definecolor{cerulean}{rgb}{0.0, 0.48, 0.65}
\definecolor{celestialblue}{rgb}{0.29, 0.59, 0.82}
\definecolor{azure(colorwheel)}{rgb}{0.0, 0.5, 1.0}
\definecolor{babyblue}{rgb}{0.54, 0.81, 0.94}
\definecolor{babyblueeyes}{rgb}{0.63, 0.79, 0.95}
\definecolor{ballblue}{rgb}{0.13, 0.67, 0.8}

\definecolor{asparagus}{rgb}{0.53, 0.66, 0.42}
\definecolor{ao(english)}{rgb}{0.0, 0.5, 0.0}
\definecolor{applegreen}{rgb}{0.55, 0.71, 0.0}
\definecolor{armygreen}{rgb}{0.29, 0.33, 0.13}
\definecolor{gray-asparagus}{rgb}{0.27, 0.35, 0.27}
\definecolor{green(ryb)}{rgb}{0.4, 0.69, 0.2}

\definecolor{amethyst}{rgb}{0.6, 0.4, 0.8}
\definecolor{antiquefuchsia}{rgb}{0.57, 0.36, 0.51}
\definecolor{blue-violet}{rgb}{0.54, 0.17, 0.89}
\definecolor{brightlavender}{rgb}{0.75, 0.58, 0.89}
\definecolor{brightube}{rgb}{0.82, 0.62, 0.91}
\definecolor{brilliantlavender}{rgb}{0.96, 0.73, 1.0}

\definecolor{amber}{rgb}{1.0, 0.75, 0.0}
\definecolor{amber(sae/ece)}{rgb}{1.0, 0.49, 0.0}
\definecolor{atomictangerine}{rgb}{1.0, 0.6, 0.4}
\definecolor{burntorange}{rgb}{0.8, 0.33, 0.0}
\definecolor{burntsienna}{rgb}{0.91, 0.45, 0.32}
\definecolor{cadmiumorange}{rgb}{0.93, 0.53, 0.18}
\definecolor{carrotorange}{rgb}{0.93, 0.57, 0.13}
\definecolor{chocolate(web)}{rgb}{0.82, 0.41, 0.12}
\definecolor{chromeyellow}{rgb}{1.0, 0.65, 0.0}
\definecolor{darkorange}{rgb}{1.0, 0.55, 0.0}
\definecolor{darktangerine}{rgb}{1.0, 0.66, 0.07}
\definecolor{deepcarrotorange}{rgb}{0.91, 0.41, 0.17}
\definecolor{deepsaffron}{rgb}{1.0, 0.6, 0.2}
\definecolor{fulvous}{rgb}{0.86, 0.52, 0.0}

\begin{document}

\title{PCF-Lift: Panoptic Lifting by\\ Probabilistic Contrastive Fusion} 



\author{Runsong Zhu\inst{1} \and
Shi Qiu\inst{1} \and
Qianyi Wu\inst{2} \and
Ka-Hei Hui\inst{1}
\and
\\
Pheng-Ann Heng\inst{1}
\and
Chi-Wing Fu\inst{1}}
\authorrunning{R.~Zhu, S.~Qiu, Q.~Wu, et al.}

\institute{The Chinese University of Hong Kong \and
Monash University}

\maketitle
\begin{abstract}
Panoptic lifting is an effective technique to address the 3D panoptic segmentation task by unprojecting 2D panoptic segmentations from multi-views to 3D scene.
However, the quality of its results largely depends on the 2D segmentations, which could be noisy and error-prone, so its performance often drops significantly for complex scenes.
In this work, we design a new pipeline coined \textbf{PCF-Lift} based on our \textbf{P}robabilis-tic \textbf{C}ontrastive \textbf{F}usion (PCF) to learn and embed probabilistic features throughout our pipeline to actively consider inaccurate segmentations and inconsistent instance IDs.
Technical-wise, we first model the probabilistic feature embeddings through multivariate Gaussian distributions.
To fuse the probabilistic features, we incorporate the probability product kernel into the contrastive loss formulation and design a cross-view constraint to enhance the feature consistency across different views.
For the inference, we introduce a new probabilistic clustering method to effectively associate prototype features with the underlying 3D object instances for the generation of consistent panoptic segmentation results.
Further, we provide a theoretical analysis to justify the superiority of the proposed probabilistic solution.
By conducting extensive experiments, our PCF-lift not only significantly outperforms the state-of-the-art methods on widely used benchmarks including the ScanNet dataset and the challenging Messy Room dataset (4.4\% improvement of scene-level PQ), but also demonstrates strong robustness when incorporating various 2D segmentation models or different levels of hand-crafted noise.
The code is publicly available at \href{https://github.com/Runsong123/PCF-Lift}{https://github.com/Runsong123/PCF-Lift}.
\end{abstract}    
\section{Introduction}
\label{sec:intro}
3D panoptic segmentation~\cite{zhou2021panoptic,sirohi2021efficientlps,milioto2020lidar,gasperini2021panoster,dahnert2021panoptic,narita2019panopticfusion,rosinol20203d} is a challenging 3D vision task, requiring the prediction of both semantic segmentation labels and instance segmentation labels.
This task enables a comprehensive understanding of 3D scenes, thus facilitating many downstream applications,~\eg, VR/AR, robotics,~\etc.

Observing that the generalizability and applicability of existing 3D panoptic segmentation methods are limited by the scarcity of 3D training data, recent studies~\cite{kobayashi2022decomposing, tschernezki2022neural,zhi2021place,siddiqui2023panoptic,bhalgat2023contrastive} suggest the idea of leveraging 2D panoptic segmentation information predicted by foundation models~\cite{cheng2022masked,zhou2022detecting,kirillov2023segment}.
Although 2D panoptic segmentation only provides image-based understanding, panoptic lifting~\cite{siddiqui2023panoptic,bhalgat2023contrastive} is an effective technique to learn implicit 3D panoptic fields from 2D panoptic predictions, supporting the generation of coherent and view-consistent panoptic segmentation across different views.

\begin{figure}[!t]
\begin{center}
\begin{overpic}[width=0.9\linewidth]{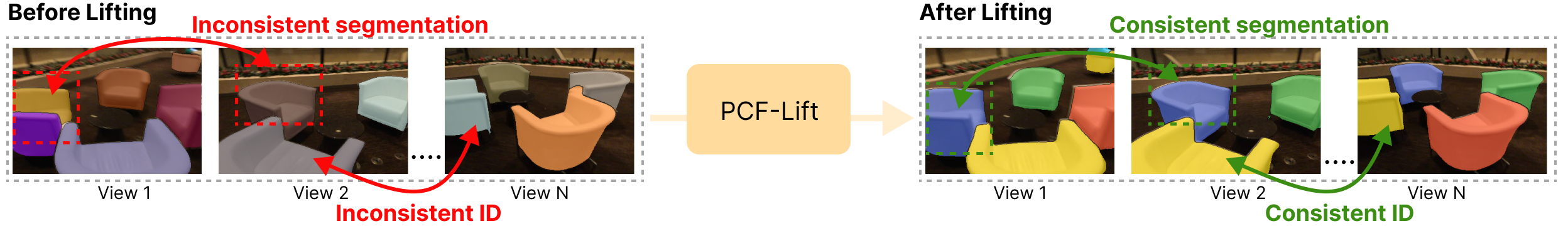}
\end{overpic}
\end{center}
%
% \vspace{-0.01cm}
\caption{
Our PCF-Lift method unprojects 2D panoptic segmentation predictions to 3D domain, facilitating the generation of consistent panoptic segmentation masks. 
For simplicity and clarity, we highlight instance segmentation masks. 
}
\label{fig:panoptic-lifting}
\end{figure}

Recent works~\cite{wang2022dm,siddiqui2023panoptic,bhalgat2023contrastive} on panoptic lifting mainly focus on solving the challenging instance-related issues, given
% the fact 
that semantic predictions can be effectively associated across different views~\cite{zhi2021place}.
In practice, as shown in Fig.~\ref{fig:panoptic-lifting}, one direct issue is \emph{inconsistent IDs} that the same 3D object is assigned to different instance IDs in two different views for 2D panoptic prediction, which complicates the straightforward fusion process.
To bypass the \emph{inconsistent IDs} issue, Panoptic Lifting~\cite{siddiqui2023panoptic} directly fits instance ID permutations during the training process to learn deterministic feature embeddings for instance representation.
Furthermore, Contrastive Lift~\cite{bhalgat2023contrastive} extends the scalability of Panoptic Lifting by using the contrastive learning technique to optimize the deterministic feature embeddings.
However, existing methods still struggle to achieve satisfactory performance on complex scenes, as they overlook another interrelated issue: \emph{inconsistent segmentation}.
Specifically, due to the presence of inaccurate segmentation, the same object can be segmented inconsistently across two views: \eg, the chair in Fig.~\ref{fig:panoptic-lifting} is segmented into two parts in ``view 1'' but as a whole in ``view 2''.
As existing methods use deterministic feature embeddings to learn 3D instance segmentation, the issue of \emph{inconsistent segmentation} inherently introduces noise to training data and thus poses a robustness challenge for the models.

In this paper, we introduce an effective probabilistic contrastive fusion solution to collectively address the two issues.
For the issue of \emph{inconsistent segmentation}, we propose to learn probabilistic feature embeddings rather than deterministic feature embeddings used in~\cite{siddiqui2023panoptic,bhalgat2023contrastive}. 
Our key insight is the development of probabilistic features, which correspond to distributions, allows for the incorporation of uncertainty modeling and enhances robustness to noise.
This leads to a stable model optimization and, ultimately, results in a more reliable representation of instance information.
Given these benefits, we develop probabilistic feature embeddings based on multivariate Gaussian distributions.
Particularly, to enable contrastive learning among different Gaussian distributions, we devise the Probability Product (PP) Kernel~\cite{jebara2004probability} to measure the probabilistic feature similarities.
Accordingly, a probabilistic clustering algorithm is introduced in the inference phase to generate consistent panoptic segmentation, using the measured probabilistic feature similarities.
For the issue of \emph{inconsistent IDs}, we devise the contrastive loss to fuse the probabilistic feature embeddings.
In addition to formulating the loss term with each single view observation as inspired by the prior work~\cite{bhalgat2023contrastive}, we introduce a novel cross-view constraint to facilitate an effective model training from segmentation results with inconsistent instance IDs.
By dynamically exploiting feature pairs from different views, our proposed constraint offers an effective formulation to further enhance the feature consistency of the same 3D object instance across multiple views, and thus improves the panoptic lifting performance.

To evaluate the effectiveness of our method, we conduct experiments on the ScanNet dataset~\cite{dai2017scannet} and the Messy Rooms dataset~\cite{bhalgat2023contrastive}.
Furthermore, we conduct experiments to demonstrate the robustness of our probabilistic method to variations in segmentation models and different levels of noise.
In addition to the experimental outcomes, our theoretical analysis from an optimization perspective demonstrates that the proposed probabilistic representation can be seen as a more flexible and generalized form compared to the previous deterministic representation~\cite{bhalgat2023contrastive}.

Our contributions are summarized as follows:
\begin{itemize}
\item We introduce a probabilistic contrastive fusion solution (PCF-Lift) to effectively unproject the 2D panoptic segmentations to the 3D domain by collectively considering the issues of \emph{inconsistent segmentation} and \emph{inconsistent IDs}.
\item To fuse the probabilistic feature embeddings modeled by multivariate Gaussian distributions, we reformulate the contrastive loss with the Probability Product kernel and propose a novel cross-view constraint to further encourage the multi-view consistency.

\item Coupled with a new probabilistic clustering algorithm, our proposed method outperforms the state-of-the-art methods consistently on the ScanNet dataset and the Messy Room dataset.

\end{itemize}

\section{Related Works}

\subsection{Traditional 2D and 3D Panoptic Segmentation}
The 2D Panoptic segmentation task was initially introduced in~\cite{kirillov2019panoptic}.
Despite the notable advancements made by subsequent works~\cite{cheng2020panoptic,porzi2019seamless,cheng2022masked,cheng2021per,zhang2021k}, 
it remains challenging to panoptically understand individual images, while avoiding inconsistent instance recognitions across different image views of the scene.
To enhance the panoptic understanding of the real world, 3D panoptic segmentation focuses on segmenting pre-computed 3D structures~\cite{zhou2021panoptic,sirohi2021efficientlps,milioto2020lidar,gasperini2021panoster} (\eg, point clouds, voxels) or performing simultaneous 3D reconstruction or segmentation from 2D images~\cite{dahnert2021panoptic,narita2019panopticfusion,rosinol20203d}.
However, its generalizability is rather limited, largely due to the significant difference in scale between 2D and 3D training data.
Given the capabilities of recent 3D reconstruction techniques~\cite{mildenhall2021nerf,chen2022tensorf,kerbl20233d} and 2D panoptic segmentation models~\cite{zhou2022detecting,cheng2022masked}, in this work, we explore 3D panoptic segmentation by leveraging multi-view images without explicit 3D input data.
\subsection{Multi-view Fusion}
Recently, researches in 3D reconstruction have made tremendous progress~\cite{mildenhall2021nerf,chen2022tensorf,barron2021mip,liu2020neural,kerbl20233d}.
Beyond novel view synthesis, we can utilize 3D reconstruction techniques as a tool to fuse 2D information like semantics and features in the 3D space~\cite{cen2024segment,ye2023gaussian,kobayashi2022decomposing,tschernezki2022neural,kerr2023lerf,fan2022nerf,hamilton2022unsupervised,zhang2023uni,nie2020total3dunderstanding}.
For example, Semantic-NeRF~\cite{zhi2021place} learns a semantic field from the 2D semantic segmentation, demonstrating the robustness and effectiveness of the semantic fusion.
Later works~\cite{kobayashi2022decomposing,tschernezki2022neural,kerr2023lerf} propose to fuse the unsupervised 2D dense feature for various segmentation and editing applications, showcasing the potential to adapt the 2D zero-shot models to 3D domain.
In this work, we study the task of panoptic fusion, integrating both instance and semantic information to obtain a comprehensive understanding of the 3D scene.
This task is significantly more challenging than the previous semantic fusion task~\cite{zhi2021place}, since we need to incorporate additional instance fusion and achieve instance label consistency.
Unlike recent methods (\eg, Panoptic-Lifting~\cite{siddiqui2023panoptic} and Contrastive Lift~\cite{bhalgat2023contrastive}), which utilize deterministic feature embeddings to represent instance information,
we develop a novel probabilistic solution, actively considering the error-prone and noisy nature of the 2D segmentations, such that we can duly enhance the effectiveness and robustness of instance fusion.
\subsection{Probabilistic Representation}
Probabilistic feature embedding is a popular tool employed in various tasks,~\eg, image generation~\cite{kingma2013auto,rezende2014stochastic}, normal estimation~\cite{bae2021estimating}, video understanding~\cite{park2022probabilistic}, point cloud understanding~\cite{cai2023uncertainty}, and prototype embeddings for few-shot detection~\cite{tang2023prototypical}.
Motivated by its high capabilities in estimating aleatoric uncertainty~\cite{der2009aleatory} and addressing data noise, 
we innovate a new panoptic-lifting pipeline for 3D panoptic segmentation by formulating new modules based on probabilistic feature embeddings, including a reformulated contrastive loss with the probability product kernel~\cite{jebara2004probability} and an effective cross-view constraint to enhance the multi-view consistency.
Further, we also design a novel probabilistic clustering algorithm to facilitate the generation of consistent panoptic segmentation.
\section{Method}
\begin{figure}[!t]
\begin{center}
\begin{overpic}[width=0.9\linewidth]{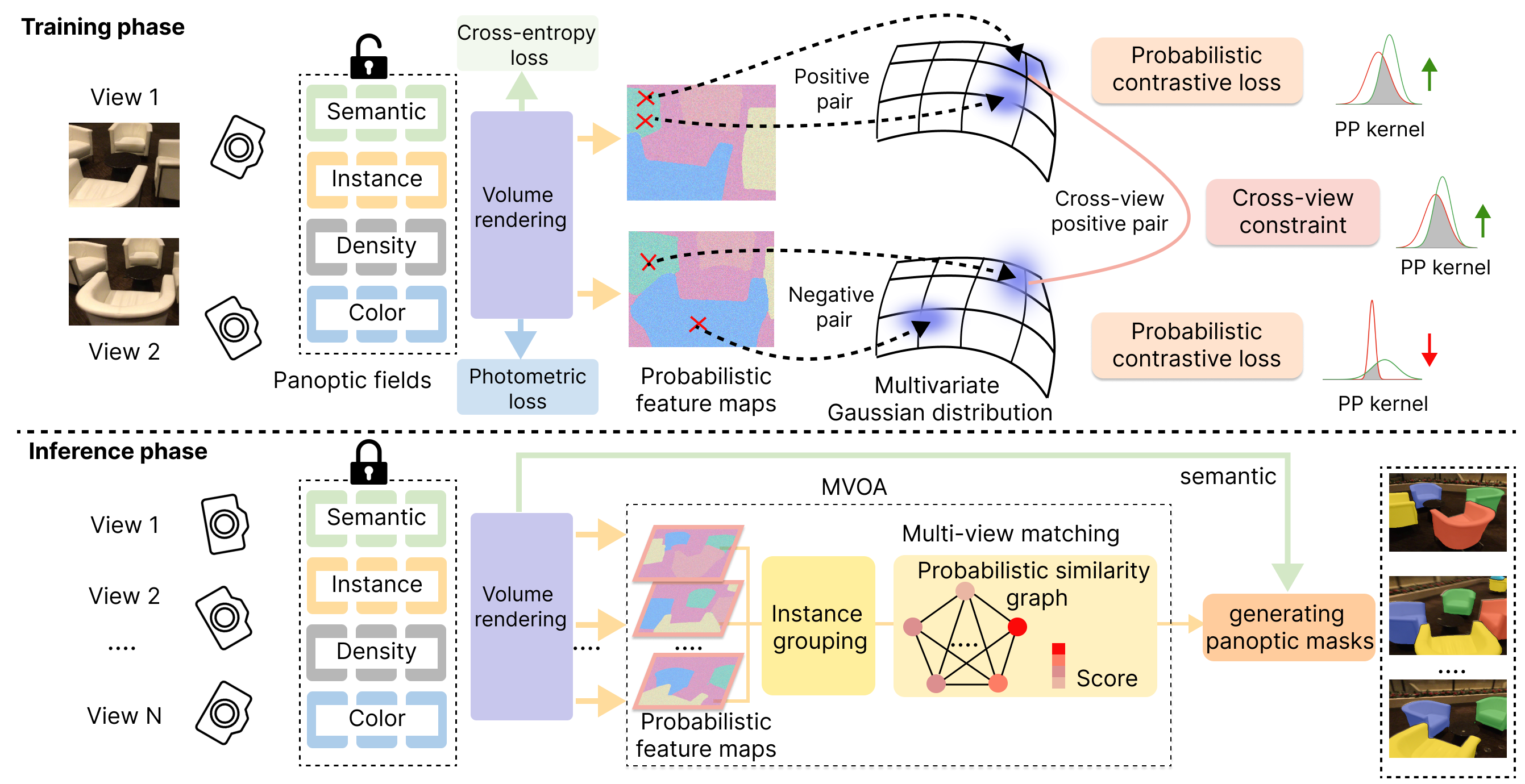}
\end{overpic}
\end{center}
\caption{
Overview of PCF-Lift.
The 3D panoptic fields include a semantic field, an instance field, a density field, and a color field.
To solve the instance-related issues, we propose to learn probabilistic feature embeddings in the instance field (see Sec.~\ref{sec:embedding}). 
During the training phase, given two camera views, we can render the probabilistic feature maps from the instance field via volume rendering.
To optimize the probabilistic instance field, we devise the probabilistic contrastive loss with Probability Product (PP) kernel~\cite{jebara2004probability}, and propose a cross-view constraint to further enhance the feature consistency from different views (see Sec.~\ref{sec:contrast-fusion}).
Similarly, we can render the semantic and color predictions, and adopt photometric loss and cross-entropy loss to optimize the semantic field, the density field, and the color field.
During the inference phase, we design a novel multi-view object association (MVOA) algorithm for the generation of consistent panoptic segmentations (see Sec.~\ref{sec:inference}).}

\label{fig:overview}
\end{figure}

Given a set of posed images $\{\mathcal{I}\}$ associated with 2D panoptic segmentation predictions (\ie, semantic masks $\{ \mathcal{H}\}$ and instance masks $\{\mathcal{K}\}$) 
generated by 2D segmentation models, our goal is to learn accurate 3D panoptic fields that can be rendered into consistent panoptic segmentation results over different views.
The overview of PCF-Lift is illustrated in Fig.~\ref{fig:overview}.
Specifically, the 3D panoptic fields include a semantic field, an instance field, a density field, and a color field.
Particularly, considering the significance of instance-related issues, we propose to learn probabilistic feature embeddings in the instances field (see Sec.~\ref{sec:embedding}).  
During the training phase, we jointly train the whole panoptic fields.
Given two camera views, we can render the probabilistic feature maps from the instance field via volume rendering~\cite{kajiya1984ray}.
To optimize the probabilistic feature embeddings, we develop the contrastive loss with the Probability Product kernel~\cite{jebara2004probability}, and propose a cross-view constraint to further enhance the feature consistency from different views (see Sec.~\ref{sec:contrast-fusion}).
Similarly, the semantic predictions and color predictions are also rendered via volume rendering, where the semantic field, the density field, and the color field are optimized by the photometric loss and cross-entropy loss~\cite{chen2022tensorf,zhi2021place,siddiqui2023panoptic,bhalgat2023contrastive}. 
In the inference phase, we design a probabilistic clustering (\ie, multi-view object association) algorithm to effectively identify the prototype features of underlying 3D object instances for the generation of consistent panoptic segmentation results across any given views (see Sec.~\ref{sec:inference}).  
\subsection{Probabilistic Feature Embeddings}
\label{sec:embedding}
To provide a robust instance representation, we propose to learn probabilistic feature embeddings in the instance field, which maps each 3D point to a random variable $\mathcal{F}$, modeled as an $N$-dimensional multivariate Gaussian distribution $\mathcal{F} \sim \mathcal{N}(\mathbf{\mu},\mathbf{\Sigma})$.
Here, $\mu$ is the mean vector, indicating the central feature values and $\mathbf{\Sigma}$ is a diagonal covariance matrix $\operatorname{diag}(\sigma^{2})$ with $\sigma^{2}=(\sigma^{(1)^{2}},\sigma^{(2)^{2}},\cdots,\sigma^{(N)^{2}})$.
Concretely, for any given query point $\mathbf{x}\in \mathbb{R}^{3}$, the instance field predicts $(\mu,\sigma^{2}) \in \mathbb{R}^{2N}$. 
Similar to the rendering of the color field, for each pixel in a given camera view, we can render its corresponding Gaussian distribution feature via volume rendering.
A significant advantage of using probabilistic features rather than deterministic features is the ability of Gaussian distributions that assign different covariance values for uncertainty modeling.
This property is crucial as it aids in reducing the impact of noise, thereby enhancing the robustness and accuracy of feature embeddings in representing complex 3D scenes.

Intuitively, the similarity between two rendered Gaussian distributions indicates whether the corresponding pixels belong to the same instance. 
To quantify the similarities between different Gaussian distributions, we employ the Probability Product (PP) kernel~\cite{jebara2004probability} $K_\rho$.
Specifically, given two Gaussian distributions
$\mathcal{F}_{i} \sim \mathcal{N}(\mathbf{\mu}_{i}, \mathbf{\Sigma}_{i})$ and $\mathcal{F}_{j} \sim \mathcal{N}(\mathbf{\mu}_{j},\mathbf{\Sigma}_{j})$, the corresponding kernel can be explicitly formulated as
\begin{equation}
\scalebox{0.85}{
$
K_\rho\left(\mathcal{F}_i, \mathcal{F}_j\right) 
=\left(\prod_{d=1}^N\left(\frac{\sigma_i^{(d)^{2}}}{\sigma_j^{(d)^{2}}}  +\frac{\sigma_j^{(d)^{2}}}{\sigma_i^{(d)^{2}}} \right) / 2\right)^{-\frac{1}{2}} \exp \left(-\sum_{d=1}^N\left(\mu_i^{(d)}-\mu_j^{(d)}\right)^{2} / 4\left(\sigma_i^{(d)^{2}}+\sigma_j^{(d) ^{2}}\right)\right).$}
\label{eq:kkp}
\end{equation}
Note that the results produced by the PP kernel fall within the range of 0 (minimal similarity) to 1 (maximal similarity), the same as the Radial Basis Function (RBF) kernel used in the deterministic method~\cite{bhalgat2023contrastive}. 
\subsection{Training Probabilistic Feature Embeddings}
\label{sec:contrast-fusion}
\subsubsection{Probabilistic Contrastive Loss.}
During the training phase, we leverage the contrastive learning technique to optimize the probabilistic feature embeddings.
Specifically, the contrastive loss used in~\cite{chen2020simple,le2020contrastive,xie2020pointcontrast,newell2017associative} tends to increase the predicted feature similarities of predefined positive pairs, while decreasing the similarities of negative pairs for training.
In our work, a positive pair is defined as two pixels belonging to the same instance, while a negative pair denotes two pixels of different instances within a single image, following the previous method~\cite{bhalgat2023contrastive}.
Since we basically learn probabilistic features, the PP kernel is exploited to measure the features similarity.
In general, the loss can be formulated as
\begin{equation}
\mathcal{L}_{\text{pixel-contra}} =- \frac{1}{|\Omega|}\sum_{u \in \Omega} \log \frac{\sum_{u^{\prime} \in \Omega} \mathbf{1}_{\left(u,{u^{\prime}}\right)} \exp \left(K_{\rho}\left(\mathcal{F}_u, {\mathcal{F}}_{u^{\prime}} \right)\right)}{\sum_{u^{\prime} \in \Omega} \exp \left(K_{\rho}\left(\mathcal{F}_u, {\mathcal{F}}_{u^{\prime}}\right)\right)},
\label{eq:contr}
\end{equation}

\begin{equation}
\mathcal{L}_{\text{concen}} = - \frac{1}{|\Omega|} \sum_{u \in \Omega} \log \left( K_{\rho}\left(\mathcal{F}_u,\frac{\sum_{u^{\prime} \in \Omega} \mathbf{1}_{{\left(u,{u^{\prime}}\right)}} {\mathcal{F}}_{u^{\prime}}}{\sum_{u^{\prime} \in \Omega} \mathbf{1}_{{\left(u,{u^{\prime}}\right)}}}\right) \right),
\label{eq:contr}
\end{equation}

\begin{equation}
\text{and}~ \mathcal{L}_{\text{contra}}= \mathcal{L}_{\text{pixel-contra}}  + \mathcal{L}_{\text{concen}},
\label{eq:contr}
\end{equation}
where $\mathcal{L}_{\text{pixel-contra}}$ is the pixel-wise contrastive loss, $\mathcal{L}_{\text{concen}}$ is the concentrate loss term, $\bbone$ is the indicator function of positive pairs, $\Omega$ is the set of pixel sample, $\mathcal{F}_{u}$ and $\mathcal{F}_{u^{\prime}}$ are the rendered features for pixels $u$ and $u^{\prime} \in \Omega$ via volume rendering, respectively.
Different from previous method~\cite{bhalgat2023contrastive}, we devise probabilistic similarity kernels to calculate a more effective contrastive loss for model training.
By optimizing the loss Eq.~\eqref{eq:contr}, we can learn a more expressive 3D instance field for robust feature representations, as further analyzed in Sec.~\ref{sec:relation}.
To avoid the neural network from generating large covariances everywhere, we use an additional regularization term, $\mathcal{L}_{\text{reg}}:=\log(\prod_{d=1}^N\sigma^{(d)^{2}})$, to penalize the large covariances.
\subsubsection{Cross-view Constraint.}
To better train the probabilistic feature embeddings, we propose a cross-view constraint to enhance feature consistency for the same object across different views. 
Given the rendered learned probabilistic feature embedding sets $\{\mathcal{F}^{m}\}$ and $\{\mathcal{F}^{n}\}$ for two different views, we devise the PP kernel $K_\rho$ and a predefined threshold $\tau$ to collect positive pairs $\mathcal{P}$ that belong to the same object: 
$\mathcal{P}=\left\{\left(\mathcal{F}_r, \mathcal{F}_s\right) \mid K_\rho\left(\mathcal{F}_r, \mathcal{F}_s\right)>\tau, {\mathcal{F}_r} \in \{\mathcal{F}^{m}\}, {\mathcal{F}_s} \in \{\mathcal{F}^{n}\} \right\}$.
Moreover, as we expect to maximize the similarities between the features that belong to the same 3D object in different views, the cross-view constraint is defined as:
\begin{equation}
\mathcal{L}_{\text{cross}} =  - \frac{1}{|\mathcal{P}|}\sum_{\left(\mathcal{F}_r, \mathcal{F}_s\right) \in \mathcal{P}} \log \left( K_\rho(\mathcal{F}_r, \mathcal{F}_s)\right).
\label{eq:cross}
\end{equation}
In practice, we use the cross-view constraint only in a later optimization stage (\ie, the last few epochs), where the feature space is sufficiently meaningful to provide reliable positive pairs. 
The overall loss is formulated as
\begin{equation}
\mathcal{L}= \mathcal{L}_{\text{contra}}  + w_\text{cross}\mathcal{L}_{\text{cross}} + w_\text{reg}\mathcal{L}_{\text{reg}},
\label{eq:contr_pro}
\end{equation}
where 
% \st{the $w_3, w_4$ is a}
$w_\text{cross}$ and $w_\text{reg}$ are weight hyper-parameters.
\subsection{Multi-view Object Association}
\label{sec:inference}
During the inference phase, to facilitate consistent panoptic segmentations, we introduce a novel clustering algorithm, named the multi-view object association (MVOA) algorithm.
The algorithm aims to extract the prototype feature set from the learned instance field for the assignment of instance labels.
In general, through volume rendering, we obtain a large-scale rendered per-pixel probabilistic feature set $\{\mathcal{F}\}$ of the training views, as the input of our algorithm.
Considering the efficiency issue, we first conduct an instance grouping operation to gather a smaller feature set $\mathcal{C}$, with the help of inconsistent instance segmentation masks $\{\mathcal{K}\}$ obtained from training views. 
Then, we construct the probabilistic similarity graph based on $\mathcal{C}$ and design a multi-view matching process to collect the prototype feature set $\mathcal{D}$.
The details of instance grouping and multi-view matching operations are introduced as follows.
\subsubsection{Instance Grouping.}
We group the feature set $\{\mathcal{F}\}$ from each individual view.
For the view of $l$-th image, we collect the pixels that belong to instance $p$ into the set ${\{i:\mathcal{K}_{l}^{i}=p\}}$, and the corresponding probabilistic feature set is denoted as $\{{\mathcal{F}^{i}_{l}:\mathcal{K}_{l}^{i}=p\}}$.
Then, we group $\{{\mathcal{F}^{i}_{l}:\mathcal{K}_{l}^{i}=p\}}$ to a single probabilistic feature by calculating the average:
$
\mathcal{C}_{l}^{p}=\sum_{i} \mathcal{F}^{i}_{l} / {\left|\{i:\mathcal{K}_{l}^{i}=p\}\right|}.
$
Concurrently, we model the score quantity $\mathcal{S}_{l}^{p}$ for $\mathcal{C}_{l}^{p}$ as an indicator of feature concentration, which aids in the subsequent multi-view matching process.
The indicator function $\Phi(\mathcal{C}_{l}^{p})$ averages the PP kernel similarities between $\mathcal{C}_{l}^{p}$ and all features in $\{{\mathcal{F}^{i}_{l}:\mathcal{K}_{l}^{i}=p\}}$.
Mathematically, $\Phi(\mathcal{C}_{l}^{p})$ follows: 
$\mathcal{S}_{l}^{p} = \Phi(\mathcal{C}_{l}^{p}) = {\sum_{\left\{i: \mathcal{K}_{l}^{i}=p\right\}}  K_{\rho}(\mathcal{C}_{l}^{p}, \mathcal{F}^{i}_{l})}/{{\left|\{i:\mathcal{K}_{l}^{i}=p\}\right|}}.$
By applying this across all observed images, we obtain the grouped feature set $\mathcal{C}$ and the corresponding score set $\mathcal{S}$.
\subsubsection{Multi-view Matching.}
To extract the structural relationships among the features, we construct an unoriented similarity graph $G = (\mathcal{C}, E)$, where the grouped feature set $\mathcal{C}$ represents the nodes and $E$ encompasses the edges that indicate quantitative feature similarities measured by the PP Kernel:
$
{E}_{<g,h>} = {E}_{<h,g>} =  K_{\rho}(\mathcal{C}_{g}, \mathcal{C}_{h}),
$
where $\mathcal{C}_{g}, \mathcal{C}_{h} \in \mathcal{C}$.
Then, we extract the prototype feature set $\mathcal{D}$ from the graph $G$ and score set $\mathcal{S}$ by a greedy procedure, which is akin to the classical greedy algorithm known as non-maximum suppression (NMS). The pseudo-code of our proposed MVOA algorithm is presented in Algo.~\ref{alg:nms}.
In practice, the selection of hyper-parameter $\mathcal{T}$ is based on the average similarity computed across grouped feature pairs identified within each view.
\subsubsection{Generating Panoptic Segmentation Masks.}
We first follow the MVOA algorithm to collect the prototype feature set $\mathcal{D}$. 
For each view, including the novel view, we then generate the semantic labels from the learned semantic field to differentiate between the background and foreground.
Coupled with a feature map rendered from the probabilistic instance field, for each foreground pixel, we finally determine its corresponding instance label by assigning the candidate index from the set $\mathcal{D}$ that exhibits the highest similarity to the rendered feature.
In practice, the MVOA algorithm needs to be conducted only once, while the extracted prototype features $\mathcal{D}$ will be used for generating the panoptic segmentation masks of all test views.

\begin{algorithm}
\caption{Multi-view object association algorithm (MVOA)}
\label{alg:nms}
\KwData{Inconsistent instance mask $\{\mathcal{K}\}$,input feature set $\{\mathcal{F}\}$,  Threshold $\mathcal{T}$}
\KwResult{Prototype features set $\mathcal{D}$ }
\SetKwFunction{getindex}{get\_index}
\SetKwFunction{mean}{mean}
\centering
\resizebox{0.9\textwidth}{!}{%
\begin{minipage}{0.45\linewidth}

  \tcp{Instance Grouping}
  ${\mathcal{C}} = \{\},$
$\mathcal{S} = \{\}$

\For{$\mathcal{K}_{l}$ in $\{\mathcal{K}\}$}{
    % Process the item\;
    $\text{IDs}$ = \{unique\_ID($\mathcal{K}_{l}$)\}
    
    \For{$\text{p}$ in $\text{IDs}$}{
        % \tcp{This is a comment}
        $\mathcal{C}_{l}^{p}=\sum_{i} \mathcal{F}^{i}_{l} / {\left|\{i:\mathcal{K}_{l}^{i}=p\}\right|}$\\

        $\mathcal{S}_{l}^{p}= \Phi(\mathcal{C}_{l}^{p})$\\
        \\
        $\mathcal{C} \gets \mathcal{C} \cup \mathcal{C}_{l}^{p}$;
        $\mathcal{S} \gets \mathcal{S} \cup \mathcal{S}_{l}^{p}$;
    }
}
\end{minipage}
\hfill
\begin{minipage}{0.45\linewidth}
% \small
  \tcp{Multi-view matching}
  $\mathcal{D} \gets \{\}$

\While{$\mathcal{S} \neq empty $}{
    $m \gets \text{argmax}~\mathcal{S}$\\
    % $[\mathcal{F}, \Sigma] = \C_{m}$\\
    $\mathcal{D} \gets \mathcal{D} \cup \mathcal{C}_{m}$\\
    $\mathcal{C} \gets \mathcal{C} -  \mathcal{C}_{m}$; $\mathcal{S} \gets \mathcal{S} - \mathcal{S}_{m}$ \\
    \For{$\mathcal{C}_{i}$ in $\mathcal{C}$}{
        
        \If{$K_{\rho}(\mathcal{C}_{m}, \mathcal{C}_{i})   \geq  ~\mathcal{T}$}{

       $\mathcal{C} \gets \mathcal{C} - \mathcal{C}_{i}$; $\mathcal{S} \gets \mathcal{S} - \mathcal{S}_{i}$
        
        }
    }
}
\end{minipage}
}

\end{algorithm}
\subsection{Theoretical Analysis}
\label{sec:relation}

\begin{figure*}[!h]
\begin{center}
\begin{overpic}[width=0.9\linewidth]{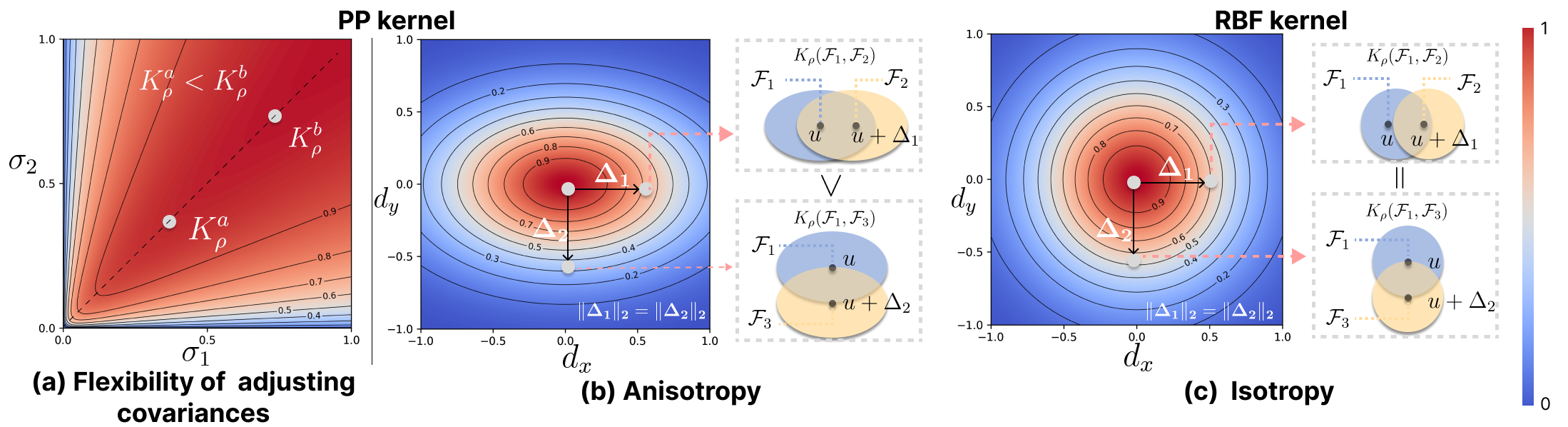}
\end{overpic}
\end{center}
\caption{
(a) Flexibility of adjusting covariances. Contour plot of the PP kernel similarity for two Gaussians with different covariance values ($\sigma_{1}$ and $\sigma_{2}$) and fixed mean values in a 1-dimensional case. 
(b) Anisotropy. Contour plot of the PP kernel similarity for two Gaussians with different Gaussian mean offsets ($d_{x}$ and $d_{y}$) and fixed covariances in a 2-dimensional case. 
(c) Isotropy. Contour plot of the RBF kernel similarity for two deterministic features with different offsets ($d_{x}$ and $d_{y}$) in a 2-dimensional case.
}
\label{fig:PPK}
\end{figure*}
\begin{prop}
    If the covariances of the given Gaussian distributions are isotropic and fixed, \ie, $\Sigma_{i}= \Sigma_{j}= \sigma \textbf{I}$, where $\sigma$ is a constant scalar, the probability product kernel can be simplified to an RBF kernel. 
    \label{rbf}
\end{prop}

The primary distinction between our probabilistic method and the prior deterministic method~\cite{bhalgat2023contrastive} is the choice of a similarity kernel. 
Specifically, our probabilistic method exploits the PP kernel, which offers higher flexibility and stronger expressive capability by adjusting the Gaussian mean and covariance compared to the RBF kernel employed in the deterministic method, as demonstrated in Fig~\ref{fig:PPK}.
The RBF kernel can be considered as a degenerate PP kernel when applied to a Gaussian distribution with an isotropic and fixed covariance, as stated in Corollary~\ref{rbf}.
From this perspective, the proposed probabilistic method is a more general framework, while the deterministic method is a subclass within the broader probabilistic paradigm.
The experimental results also verify the effectiveness of our probabilistic solution in the ablation studies of Sec.~\ref{sec:ablation}.
\section{Experiment}

\subsection{Experimental Settings}
\subsubsection{Implementation Details.}
For fair comparison, we adopt the same architecture of TensoRF~\cite{chen2022tensorf} together with the same layer parameters used in previous works~\cite{siddiqui2023panoptic,bhalgat2023contrastive}.
Particularly, our instance field is constructed following the slow-fast architecture of Contrastive Lift~\cite{bhalgat2023contrastive}, where a shallow 5-layer MLP is applied to predict an probabilistic feature embedding for a 3D coordinate input.
In practice, the dimension $N$ for the probabilistic feature embedding is 3 in our experiments.
For the training of the instance field, we sample rays from two different views in the last few epochs for optimization (Sec.~\ref{sec:contrast-fusion}), using the cross-view constraint $\mathcal{L}_{\text{cross}}$ with a predefined threshold value $\tau=0.9$ and weight $w_{\text{cross}}=0.05$; while in other training epochs, we set the weight $w_{\text{cross}}$ to 0 and only sample the rays from single views.
Besides, $w_{\text{reg}}$ in Eq.~\eqref{eq:contr_pro} is set to 0.001 throughout the whole training process.
For the training of the color, density and semantic fields, we adopt the same loss terms and training strategies from previous works~\cite{siddiqui2023panoptic,bhalgat2023contrastive}.
More implementation details are provided in the supplementary material.
\subsubsection{Metrics.} 
Since our method is proposed to solve the panoptic-lifting task, we employ scene-level Panoptic Quality ($\text{PQ}^{\text{scene}}$) for evaluation, which was introduced in Panoptic Lifting~\cite{siddiqui2023panoptic} and widely used in the related works~\cite{bhalgat2023contrastive,siddiqui2023panoptic}.
Unlike the standard PQ metric~\cite{kirillov2019panoptic}, this metric particularly considers the consistency of instance IDs across multiple views.
By merging predictions and ground truths with consistent instance IDs into subsets for $\text{PQ}^{\text{scene}}$-based evaluation, we determine a match of subset pair when the intersection over union (IoU) exceeds 0.5, in line with prior baselines~\cite{siddiqui2023panoptic,bhalgat2023contrastive}.
Since $\text{PQ}^{\text{scene}}$ is a product of scene-level segmentation quality ($\text{SQ}^{\text{scene}}$) and recognition quality ($\text{RQ}^{\text{scene}}$), we also provide the $\text{SQ}^{\text{scene}}$ and  $\text{RQ}^{\text{scene}}$ metrics for more detailed comparisons.
\subsubsection{Baselines.}
We mainly compare the proposed method with the current state-of-the-art methods that target lifting 2D panoptic predictions to 3D, \ie, Contrastive Lift~\cite{bhalgat2023contrastive} and Panoptic Lifting~\cite{siddiqui2023panoptic}.
Moreover, we compare our method with the recent NeRF-based 3D panoptic segmentation approaches, \ie, Panoptic Neural Fields~\cite{kundu2022panoptic} (PNF) and DM-NeRF~\cite{wang2022dm}.

\subsubsection{Datasets.} 
Following the previous works~\cite{siddiqui2023panoptic,bhalgat2023contrastive}, we conduct experiments on two public datasets, the ScanNet dataset~\cite{dai2017scannet} and the Messy Room dataset~\cite{bhalgat2023contrastive}, for both quantitative and qualitative evaluations.
For fair comparisons on ScanNet, following the other state-of-the-art approaches~\cite{siddiqui2023panoptic,bhalgat2023contrastive}, we adopt Mask2Former (M2F)~\cite{cheng2022masked} to generate 2D panoptic segmentation predictions, coupled with the same protocol in~\cite{siddiqui2023panoptic} that maps the COCO~\cite{lin2014microsoft} vocabulary to 21 classes.
For the experiments on Messy Rooms, we use the LVIS~\cite{gupta2019lvis} vocabulary and the Detic~\cite{zhou2022detecting} 2D panoptic segmentations, which are also utilized in Contrastive Lift~\cite{bhalgat2023contrastive}. 

\subsection{Main Experiments}
\begin{table}[htbp]
  
    \centering
    \caption{Results on the ScanNet dataset.
We report the PQ$^\text{scene}$, SQ$^\text{scene}$, and RQ$^\text{scene}$ metrics.
Since Contrastive Lift~\cite{bhalgat2023contrastive} does not report the performance using the SQ$^\text{scene}$ and RQ$^\text{scene}$ metrics, we apply the officially-released pre-trained model and re-run the clustering algorithm to obtain the values reported in this table.
For the other metric values, we directly report the ones in previous papers~\cite{bhalgat2023contrastive,siddiqui2023panoptic}.
}
\setlength{\tabcolsep}{6pt}{
    \resizebox{0.9\linewidth}{!}{
    \begin{tabular}{lccccc}
\toprule
Method& Venue & Type  & SQ$^\text{scene}$(\%) & RQ$^\text{scene}$(\%) & \textbf{PQ$^\text{scene}$(\%)} \\
\midrule
DM-NeRF~\cite{wang2022dm}& ICLR'23 & 3D panoptic segmentation   & 53.3& 46.1 & 41.7 \\
PNF~\cite{kundu2022panoptic}& CVPR'22& 3D panoptic segmentation    & 63.0 & 50.7 & 48.3\\
PNF~\cite{kundu2022panoptic} + GT BBoxes & CVPR'22 & 3D panoptic segmentation  &70.0 &55.9 & 54.3\\
\midrule
Panoptic Lifting~\cite{siddiqui2023panoptic} & CVPR'23 & 2D panoptic Lifting & 73.5 & \cellcolor{yellow!25} 65.0 & 58.9\\
Contrastive Lift~\cite{bhalgat2023contrastive} & NeurIPS'23 &2D panoptic Lifting & \cellcolor{yellow!25} 75.7 & 63.6& 
\cellcolor{yellow!25} 62.0\\
Ours  & - & 2D panoptic Lifting  & \cellcolor{red!25}78.5 & \cellcolor{red!25}65.4 &  \cellcolor{red!25}63.5 \\
\bottomrule
\end{tabular}
}
}

\label{tab:scannet-res}

\end{table}
{
\setlength{\tabcolsep}{0.04em}

\begin{table*}[!h]
\centering
\caption{
Results on the Messy Rooms dataset~\cite{bhalgat2023contrastive}.
Following~\cite{bhalgat2023contrastive}, the PQ$^\text{scene}$ metric is reported on both ``old room'' and ``large corridor'' environments with an increasing number of objects in the scene ($25, 50, 100, 500$).}
\setlength{\tabcolsep}{7.5pt}{
\resizebox{0.9\columnwidth}{!}{
\begin{tabular}{lccccccccc}

\toprule
\multirow{ 2}{*}{Method/ Number} & \multicolumn{4}{c}{Old Room Environment (\%)} & \multicolumn{4}{c}{Large Corridor Environment(\%) } &  \multirow{ 2}{*}{\textbf{Mean(\%)}} \\
\cmidrule(l{4pt}r{4pt}){2-5} \cmidrule(l{4pt}r{4pt}){6-9} 
& $25$  & $50$  & $100$  & $500$  & $25$  & $50$  & $100$  & $500$ &    \\
\midrule
 Panoptic Lifting~\cite{siddiqui2023panoptic} & 73.2 & 69.9 & 64.3 & 51.0   & 65.5 & 71.0   & 61.8 & 49.0  & 63.2 \\
 Contrastive Lift ~\cite{bhalgat2023contrastive}  & \cellcolor{yellow!25}78.9 & \cellcolor{yellow!25}75.8 & \cellcolor{yellow!25}69.1 & \cellcolor{yellow!25}55.0   & \cellcolor{yellow!25}76.5 & \cellcolor{yellow!25}75.5 & \cellcolor{yellow!25}68.7 & \cellcolor{yellow!25}52.5   & \cellcolor{yellow!25} 69.0\\
 Ours & \cellcolor{red!25}80.9  & \cellcolor{red!25}78.3    & \cellcolor{red!25}74.8  & \cellcolor{red!25}60.3  & \cellcolor{red!25}81.0    & \cellcolor{red!25}79.4  & \cellcolor{red!25}74.0  & \cellcolor{red!25}58.8 & \cellcolor{red!25} 73.4 \\

\bottomrule

\end{tabular}
}
}

\label{tab:MOS-results}
\end{table*}
}
\subsubsection{ScanNet Dataset.}
\begin{figure*}[!ht]
\begin{center}
\begin{overpic}[width=0.7\linewidth]{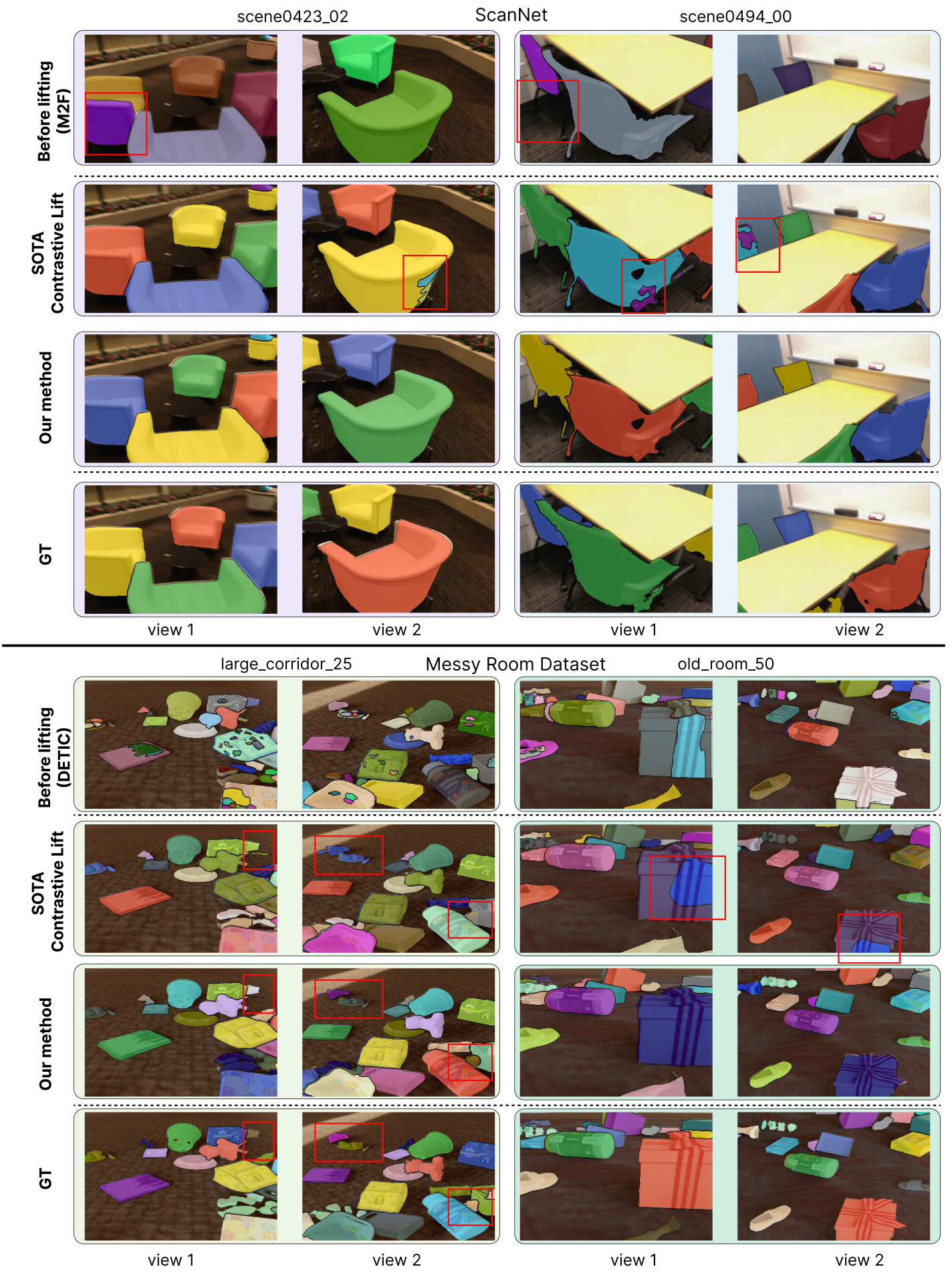}
\end{overpic}
\end{center}
\caption{
Visual comparison of the latest state-of-the-art method Contrastive Lift~\cite{bhalgat2023contrastive} and our method on the ScanNet~\cite{dai2017scannet} dataset and the Messy Room~\cite{bhalgat2023contrastive} dataset.
}
\label{fig:scannet}
\end{figure*}

To verify the performance on real data, we conduct experiments on the ScanNet dataset~\cite{dai2017scannet} of 12 scenes.
Quantitative comparisons provided in Tab.~\ref{tab:scannet-res} demonstrate that our method consistently outperforms previous baselines, including both the 3D panoptic segmentation approaches~\cite{kundu2022panoptic,wang2022dm} and the recent state-of-the-art methods for lifting 2D panoptic segmentation~\cite{bhalgat2023contrastive,siddiqui2023panoptic}.
Moreover, visual comparisons between our method and the latest state-of-the-art method~\cite{bhalgat2023contrastive} are presented in Fig.~\ref{fig:scannet}, which exhibits our method's capability of achieving a consistent and accurate 3D panoptic segmentation.

\subsubsection{Messy Rooms Dataset.}
We also conduct experiments on the challenging Messy Room dataset~\cite{bhalgat2023contrastive}, which is provided by Contrastive Lift~\cite{bhalgat2023contrastive}, containing up to 500 objects in each scene.
We present the quantitative results in Tab.~\ref{tab:MOS-results}, showing that our method (73.4\%) achieves significant improvements in terms of the mean PQ$^\text{scene}$ results compared to the current state-of-the-art methods of Contrastive Lift~\cite{bhalgat2023contrastive} (69.0\%) and Panoptic Lifting~\cite{siddiqui2023panoptic} (63.2\%).
The overall performance highlights the advantages of our probabilistic approach over the previous deterministic methods, particularly in segmenting complex scenes with hundreds of objects.
Further, our method achieves a mean SQ$^\text{scene}$ of 82.2\% and a mean RQ$^\text{scene}$ of 86.9\%, surpassing the SQ$^\text{scene}$ of 77.7\% and RQ$^\text{scene}$ of 86.6\% obtained by Contrastive Lift~\cite{bhalgat2023contrastive}.
Moreover, the visual comparisons in Fig.~\ref{fig:scannet} further indicate that our method can accurately segment the small objects in the red frames, whereas Contrastive Lift~\cite{bhalgat2023contrastive} struggles to distinguish such instances.

\subsection{Ablation Study}
\label{sec:ablation}
\subsubsection{Effectiveness of Each Component.}
\begin{table*}[h!]
\centering
\caption{Ablation study on the Messy Room dataset~\cite{bhalgat2023contrastive}.
The model (a) corresponds to Contrastive Lift~\cite{bhalgat2023contrastive} and the model (f) corresponds to our full method (PCF).}
\resizebox{0.8\linewidth}{!}{
\begin{tabular}{llcccc}
\toprule
 Model &  Feature space & Clustering  & SQ$^\text{scene}$(\%) & RQ$^\text{scene}$ (\%) & PQ$^\text{scene}$(\%)\\ \midrule
(a) & Deterministic~\cite{bhalgat2023contrastive}   & HDBSCAN~\cite{mcinnes2017hdbscan}  &  77.7 & 86.6 & 69.0\\

(b) & Deterministic~\cite{bhalgat2023contrastive}  & MVOA  &79.3 & 86.2 & 70.4 \\
\hline
(c) & Learned Gaussian distribution & HDBSCAN~\cite{mcinnes2017hdbscan}  & 78.0 &  86.6& 69.6 \\
(d) & Learned Gaussian distribution & MVOA  & 81.3 &  86.8 & 72.3\\
(e) & Learned Gaussian distribution (+ Cross-view constraint) & HDBSCAN~\cite{mcinnes2017hdbscan}  &  78.8 & 86.9  & 70.4\\
(f) & Learned Gaussian distribution (+ Cross-view constraint) & MVOA  &  82.2 &  86.9 & 73.4\\
\bottomrule
\end{tabular}
}
\label{tab:ablation}
\end{table*}
We conduct an ablation study on the Messy Room dataset~\cite{bhalgat2023contrastive}, which contains more challenging scenes with hundreds of instances.
As shown in Tab.~\ref{tab:ablation} (b) and (d), the proposed probabilistic feature embeddings greatly benefit our method, while replacing it with deterministic one leads to significant performance drop.
Furthermore, the effectiveness of our proposed cross-view constraint is verified in Tab.~\ref{tab:ablation} (e) and (f).
For the proposed multi-view object association (MVOA) algorithm, the results in Tab.~\ref{tab:ablation} (c), (d), (e), and (f) show that it particularly benefits the clustering of our proposed probabilistic representation.
Also, MVOA can be effectively employed to the deterministic method~\cite{bhalgat2023contrastive} for a performance boost, as Tab.~\ref{tab:ablation} (a) and (b) show.
% .

\subsubsection{Uncertainty Analysis.}
\begin{figure*}[!t]
\begin{center}
\begin{overpic}[width=0.8\linewidth]{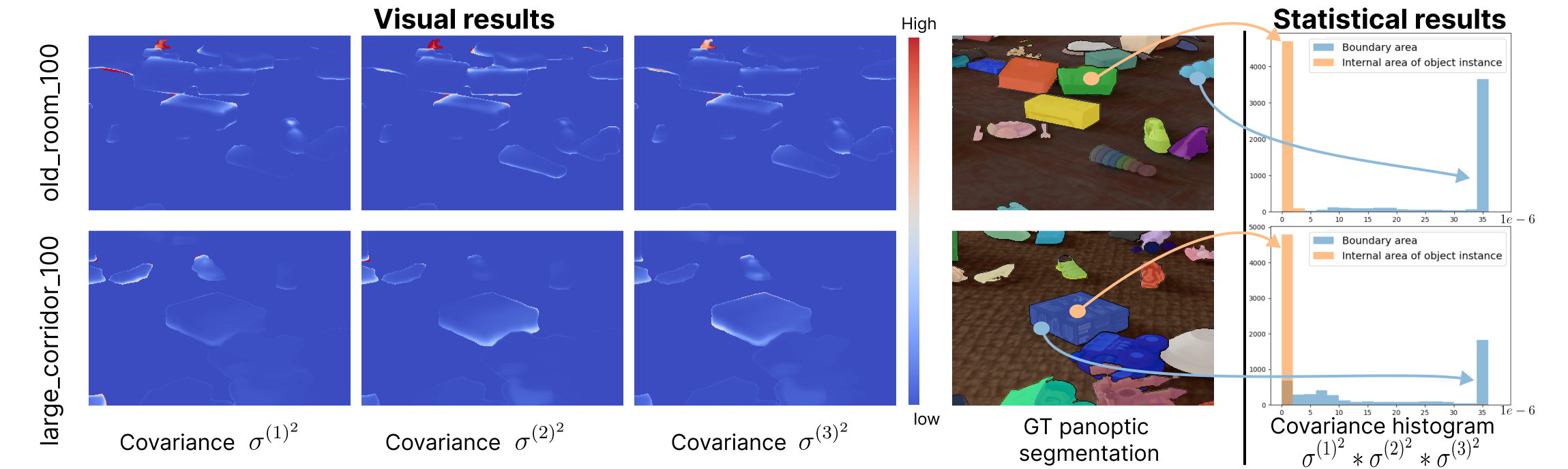}
\end{overpic}
\end{center}
\caption{
The visualization results of learned covariance components and the statistical results of covariances in two scenes of the Messy Room dataset~\cite{bhalgat2023contrastive}.
For the histograms, the horizontal axis denotes the range of covariance magnitudes, while the vertical axis corresponds to the frequency statistics for those magnitudes. We calculate and plot the covariance magnitudes ($\sigma^{(1)^{2}}*\sigma^{(2)^{2}}*\sigma^{(3)^{2}}$) of two distance image regions, the boundary areas and the internal areas of object instances, across all observed views.
}
\label{fig:uncertainty}

\end{figure*}

To verify whether our method could provide meaningful modeling for uncertainty, we provide the rendered uncertainty maps on two scenes of the Messy Room dataset~\cite{bhalgat2023contrastive} in Fig.~\ref{fig:uncertainty}.
The figure clearly illustrates that the regions of high covariance are mainly located near the boundaries of the object instances, due to the \emph{inconsistent segmentation} issue that leaves severe ambiguity around the instance boundaries.
Moreover, we statistically analyze the learned covariances in two distinct image regions, namely the boundary areas and the internal areas of object instances, across all observed views. 
These results further demonstrate our method's ability to model high uncertainty in areas where the instance boundaries are ambiguous.
We provide more details in the supplementary material.

\setcounter{footnote}{0}
\subsection{Robustness Experiments}
\label{sec:robust}
\begin{figure*}[!h]
\begin{center}
\begin{overpic}[width=0.7\linewidth]{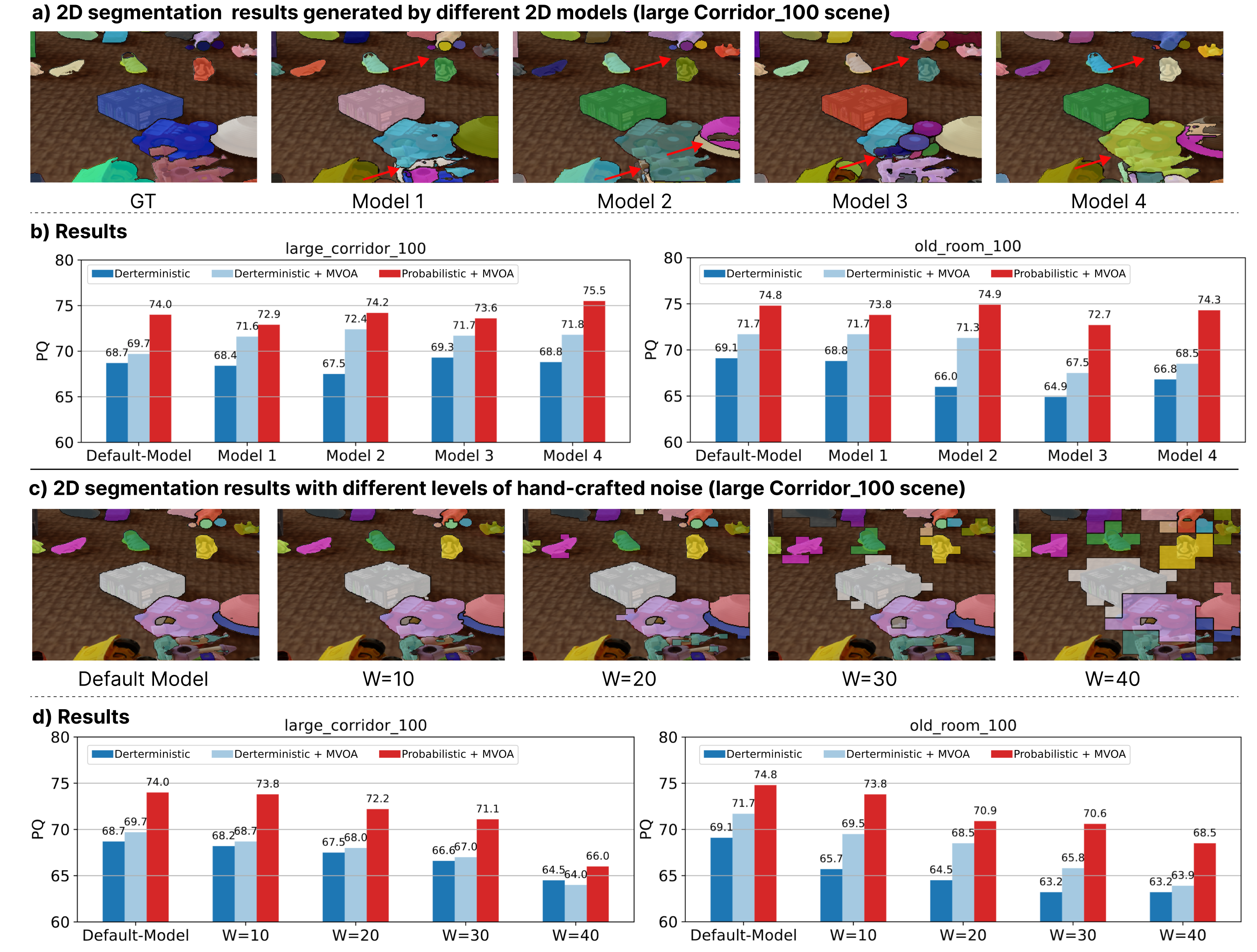}
\end{overpic}
\end{center}
\vspace{-0.2cm}
\caption{Quantitative comparisons when using different 2D models or adding various levels of hand-crafted noise.
We compare our probabilistic method (``Probabilistic $+$ MVOA'' in red) with the deterministic method ~\cite{bhalgat2023contrastive} (``Deterministic'' in dark blue) and a variant of deterministic method using our proposed MVOA algorithm (``Deterministic $+$ MVOA'' in light blue).
Besides, the four tested models are from the model zoo of Detic~\cite{zhou2022detecting}\protect\footnotemark, and the two tested scenes are from the Messy Room dataset~\cite{bhalgat2023contrastive}.
}
\label{fig:robustness}
\end{figure*}
\subsubsection{2D Backbones.}
In practice, the quality of panoptic segmentation generated by different 2D models may vary a lot.
To study the robustness of our probabilistic method and the deterministic method~\cite{bhalgat2023contrastive} when incorporating with different 2D models, we select four different models from the model zoo of Detic~\cite{bhalgat2023contrastive} and utilize the LVIS~\cite{gupta2019lvis} vocabulary to generate the 2D panoptic segmentation predictions for lifting, as shown in Fig.~\ref{fig:robustness} (a).
The quantitative results are presented in Fig.~\ref{fig:robustness} (b), where our probabilistic methods consistently outperform the baseline method by a large margin.
Particularly, we observe that the proposed MVOA algorithm can consistently boost the performance of a deterministic method~\cite{bhalgat2023contrastive}.
\subsubsection{Hand-crafted Noise.}
We study the robustness by adding hand-crafted noise.
Specifically, for all given panoptic segmentation masks,
we randomly select 100 pixels as anchors.
Then, for each anchor, we randomly choose a pixel within a $W*W$ window centered on the anchor and assign its instance ID to all pixels within this window, to simulate inaccurate segmentation predictions around object boundaries.
As shown in Fig.~\ref{fig:robustness} (c), as W gradually increases, the instance boundaries are continuously blurred.
Our probabilistic approach consistently achieves the best performance as presented in Fig.~\ref{fig:robustness} (d) and the results demonstrate that the proposed MVOA algorithm benefits the deterministic method~\cite{bhalgat2023contrastive}.
\footnotetext{We use the official pre-trained models provided in~\href{https://github.com/facebookresearch/Detic/tree/main}{https://github.com/facebookresearch/Detic/tree/main}. Please refer to our supplementary material for the details of the four models.}

\vspace{-0.1cm}
\section{Conclusion}
\vspace{-0.1cm}
We present PCF-Lift for the panoptic lifting task.
First, we propose to learn the probabilistic feature embeddings through a multivariate Gaussian distribution for instance representation.
For training, we reformulate the contrastive loss with the Probability Product kernel and propose a novel cross-view constraint to enhance the feature consistency across different views.
During the inference phase, we propose a novel multi-view object association algorithm to effectively identify the prototype features representing underlying 3D object instances.
We verify the effectiveness and robustness of PCF-Lift by extensive experiments.
% 

\section*{Acknowledgements} 
This work is supported by the InnoHK of the Government of Hong Kong via the Hong Kong Centre for Logistics Robotics, the Shenzhen Portion of Shenzhen-Hong Kong Science and Technology Innovation Cooperation Zone under HZQB-KCZYB-20200089, and the CUHK T Stone Robotics Institute.
\bibliographystyle{splncs04}
\bibliography{egbib}
\end{document}